 \documentclass[final,3p,times]{elsarticle}
\usepackage{amssymb}
\usepackage{algorithm}
\usepackage{algorithmic}
\usepackage{float}
\usepackage{makecell}
\usepackage{multirow}
\usepackage{adjustbox}
\usepackage{lscape} 
\usepackage{caption}
\usepackage{subcaption}
\usepackage{amsmath}
\usepackage{amsfonts}
\usepackage{sidenotes}
\usepackage{natbib}
\usepackage{pifont}
\usepackage{hyperref}
\usepackage{xcolor}
\usepackage{booktabs}
\usepackage{amsmath} 
\usepackage{mathrsfs}

\hypersetup{colorlinks=true,linkcolor=blue,citecolor=green,urlcolor=red}

\journal{XXXX}

\begin{document}

\begin{frontmatter}

\title{Revitalizing Electoral Trust: Enhancing Transparency and Efficiency through Automated Voter Counting with Machine Learning}

\author[1]{Mir Faris}
 \ead{mirfaris79@gmail.com}
 
 \author[1]{Syeda Aynul Karim}
 \ead{syedaishmee19@gmail.com}

\author[1]{Md. Juniadul Islam}
	 \ead{islammdjuniadul@gmail.com}

\address[1]{Department of Computer Science, American International University-Bangladesh, Dhaka-1229, Bangladesh}

\begin{abstract}

This study aims to investigate the feasibility of utilizing sophisticated image processing techniques for 
automated voter counting as a solution to problems related to manual vote counting during electoral 
processes. The goal of the study is to clarify how automated systems that make use of state-of-the-art 
technologies like OpenCV, CVZone and the MOG2 algorithm could significantly improve election 
operations' efficiency and transparency. The empirical results highlight the potential of automated voter 
counting to improve voting procedures and restore public faith in election results, especially in areas 
where trust is low. The study also stresses the use of strict measures, like the F1 score, to evaluate 
automated systems' accuracy in comparison to manual counting techniques in a methodical manner. This 
methodology enables a detailed comprehension of the differences in performance between automated 
and human counting techniques by providing a nuanced assessment. The incorporation of said measures 
serves to reinforce an extensive assessment structure, guaranteeing the legitimacy and dependability of 
automated voting systems inside the electoral sphere.

\end{abstract}

\iffalse
\begin{highlights}
\item Research highlight 1
\item Research highlight 2
\end{highlights}
\fi

\end{frontmatter}
\section{Introduction}
\label{sec:intro}
Elections are essential for the functioning of democratic societies. Ensuring their fairness and transparency is crucial for maintaining the trust of the electorate . However, in many parts of the world, manual vote counting has proven to be a flawed system. The time-intensive nature of manual counting introduces human error, inefficiencies, and opportunities for manipulation. These issues often culminate in a loss of public confidence, particularly in under developing nations where election outcomes are frequently contested.

In this study, It is investigated how modern technologies, particularly machine learning and image processing, can be applied to solve these issues. By automating the voter counting process, aimed to minimize human error and improve efficiency, accuracy, and transparency in electoral systems. This paper specifically focuses on Bangladesh as a case study, where electoral mistrust has become a prominent issue throughout the elections \cite{31}.

\subsection{Background}
The democratic process, as the foundation of modern societies, gives citizens the fundamental right to vote and actively participate in the choosing of their leaders. At the core of this democratic exercise lies the crucial need for accurate and efficient voter counting \cite{5}. Democracy depends on the integrity of elections, yet traditional manual vote counting is labour-intensive, and time-consuming, and it sometimes causes major delays in the reporting of results. Additionally, the potential for human error introduces a significant risk to the accuracy of the count, raising concerns about the transparency and fairness of the electoral process \cite{6}. Election results in Bangladesh and other nations have become less credible and accurate due to human vote counting. The discrepancy between election commission reports and public opinion has led to a serious crisis of trust in democracy in Bangladesh, as it has in many other places \cite{7}. Autonomous vote counting is essential for transparency and confidence since mistrust undermines democracy, particularly in Bangladesh. Automated voter counting systems, especially those incorporating image processing techniques and transparency measures like CCTV installations, hold the potential to revolutionize the electoral process\cite{8}. Technological innovations can speed up vote counting, reduce mistakes, increase transparency, simplify elections, and contribute to rebuilding public confidence in the political process. One example of this is IoT systems that combine CCTV with person recognition. A potential remedy for the problems associated with manual vote counting is automated voter counting that makes use of IoT and image processing \cite{9}\cite{10}. The use of image processing in this new method aims to capture ballot images, which could lower the time and expense of manual vote counting while enhancing accuracy, as proven in several studies. Lin et al. \cite{2} demonstrates how residual and convolutional layers are used by deep learning frameworks for video-based car counting. Another study introduces an automatic measuring and counting procedure using multiple picture analyses, streamlining the process with minimal calibrations \cite{3}. In a different context, researchers propose a new method for detecting and counting trees in aerial images using image processing techniques and machine learning algorithms \cite{4}. Additionally, Arputhamoni and Saravanan \cite{1}  introduces an intelligent online voting system uses image processing, convolutional neural networks (CNNs), and biometrics, such as fingerprint and face recognition. This study attempts to decrease voter turnout by tackling electoral fraud in India and demonstrating how voter digitization can improve Bangladeshi election accuracy by using machine learning to enable transparent and reliable automated vote counting.

\subsection{Problem Statement}
A chronic mismatch between official voter turnout and public perception poses serious difficulties to Bangladesh's democracy and creates a crisis of faith in the political process. By adding to the time and chance of human error, manual vote counting exacerbates these problems. In order to give a dependable and transparent solution, this study attempts to investigate automated voter counting using cutting-edge image processing technology. This will finally restore legitimacy and trust in the electoral system and inspire Bangladeshi voters to have newfound hope in democracy.

\subsection{ Research Motivation}
The study discusses the problems that the voting system in Bangladesh is facing, where a large discrepancy between recorded voter turnout and public opinion has resulted in a crisis of faith in free and fair elections. This confidence has been further undermined by the laborious and error-prone manual vote counting process. Thus, there is a pressing need for creative ways to improve the voting process's dependability and transparency. The goal of this research is to demonstrate the efficacy of an automated voter counting system that employs image processing techniques in lowering errors and regaining public confidence. This thesis advances the conversation on electoral reform by highlighting the function of technology in fostering integrity and transparency.

\subsection{Research questions:}
The primary research questions were:\par
\vspace{.2cm} 
RQ1:Can an automated voter counting system using machine learning and image processing serve as a reliable alternative to manual vote counting?\par
\vspace{.2cm}
RQ2:How does the adoption of such a system improve public trust in the electoral process, especially in countries like Bangladesh with significant trust deficits?\par 

\subsection{Objectives}
The primary objectives of this research are:

\subsubsection{To develop and test an automated voter counting system using advanced image processing technologies such as OpenCV, CVZone, and the MOG2 algorithm.}
\subsubsection{To compare the system’s performance in terms of accuracy, efficiency, and speed against traditional manual vote counting.}
\subsubsection{To evaluate the broader impact of the system on public trust and electoral transparency.}
\subsubsection{To explore the scalability of the system for use in national elections, especially in regions with significant political and social challenges.}
\vspace{.5cm}

\section{Literature Review}
\label{sec:LiteratureReview}
The electoral process in community totally based on the electorate system management where 
corruption on the integrity and fairness of proper vote counting issue is under concern by the citizens 
of respected countries especially Bangladesh \cite{11}. The article presents a contentious proposal for improving public confidence through the use of a computerized voting tabulation mechanism that monitors individual voter participation, with the goal of promoting fairness in elections. It emphasizes the importance of "canvassing" in counting and authenticating ballots, giving special attention to state laws that uphold openness in electoral activities. There are manual and electronic voting process while majority 
used manual and sparingly used the EVM (Electronic Voting Machine) in different city corporation 
election. As claims from the Bangladesh Nationalist parties and other political parties goes against 
EVM that likely to be rigged \cite{12}\cite{13}. 

\subsection{Traditional Voting}
As Bangladesh polling utmost relying upon manual voting, a ballot paper based system that consisting 
of controversies on electoral fraud. The voting procedures with EVM is also correlated, yet the 
Bangladesh Election Commission (BEC) stated the EVM will not be used in 2024 election due to 
financial issues to supply all the 300 constituencies \cite{15}. Typical voting 
process arises from voter’s arrival at their designated polling station till they return the special marking 
stamp to the Assistant Presiding Officer after submitting the paper in the ballot box. The voter first reaches the first polling officer to check the entry as a voter then head towards second polling officer 
who inspects the voter’s thumb with indelible ink. Then voter reaches to assistant presiding officer to 
conclude the process. The problem arises when sometimes name does not appear on the voter list, 
already ticked the list or spoilt paper issues sums up \cite{14}. Perhaps, it is sometimes hard to maintain all the voters count process and insidious 
rigging manually with the organized personnel.

\subsection{Human Detection and Count}
Machine learning technology implemented and installing the necessary equipment in situated camera 
on the purpose of counting voters at the polling booth entry. Every human body has inconsistent 
features which can be used to uniquely identify each of them using deep learning algorithms. The entry 
of the polling booths should be cautious enough to maintain the voters line and ensure the discipline 
manually. This proposes a concept to count the total inputs observation into booths and display on 
every constituency and media officially to verify the accuracy of ballot inputs. OpenCV is used as an 
open source platform along with python library CVZone, and algorithms are Haar Cascade and CNN 
deep learning ML models. 
OpenCV is stated to be an open source computer vision library considering especially complicated 
image processing and detecting objects via algorithms. It is applied in a number of real-world 
situations. It facilitates improved object recognition, gesture comprehension, and human-computer 
interaction. It is utilized, for instance, in facial recognition, camera tracking, and motion comprehension \cite{26}. It also aids with activities like measuring distances and deriving 3D information from 
multi-camera photos. To put it another way, OpenCV can be thought of as a tool that aids computers in seeing and comprehending the world around them. Examples of this kind of assistance include facial 
recognition in photos and guiding robots around their environment\cite{16}. A Python package called CVZone facilitates mediapipe operation. Because of its ability to process 
both audio and visual input, it simplifies tasks like recognizing faces, tracking hand motions, calculating body positions, and identifying facial features. To put it another way, CVZone facilitates 
the analysis of sounds and movies, which makes it simpler to perform interesting tasks like hand tracking and facial recognition in videos \cite{17}.  Thus, human detection and counting the voters 
by using this source of libraries possibly a better option as the system would need human features in 
its tasks.

\subsection{MoG2}
An algorithm to efficiently detect object by subtracting the background and reading its movement to 
count the voters individually, Mixture of Gaussian 2 (MoG2) is suitable for real time. For every pixel 
in a video clip, the MoG2 technique tracks multiple Gaussian distributions simultaneously. In contrast 
to MoG, MoG2 preserves unique density functions for every pixel, which enables it to efficiently 
handle and integrate intricate multi-modal background distributions. 
According to Mohamed et al. \cite{19}, the Mixture of Gaussians (MoG) approach proves advantageous 
in outdoor settings as it excels in distinguishing and suppressing non-stationary objects often 
considered as noise, like moving leaves or shifting sky patterns. It functions by maintaining a 
background model that is updated every time a new frame of video is captured \cite{27}. This feature 
significantly reduces its memory demands compared to non-recursive methods, enhancing its 
computational efficiency while maintaining reliability in object detection.  
The Mixture of Gaussian (MOG) approach was presented by Stauffer and Grimson as a way to describe background information pixel-by-pixel. The method of this approach was to improve the intensity and variance mean for each and every individual pixel in real time, while also changing the background continually. As MoG2 technology determines the appropriate number of Gaussian distributions for each pixel automatically. It might even choose whether or not to detect shadows with it. Because this approach can adapt to different lighting conditions, it works effectively in a variety of scenarios. MOG2 is updated version of MoG, generating similar approach with extra features \cite{18}.  By using this method, the model can adjust to changing lighting and dynamic changes in the surroundings, which helps it distinguish between various background modes that are present in the scene.

\begin{figure}[H]
    \centering
    \includegraphics[width=0.7\textwidth]{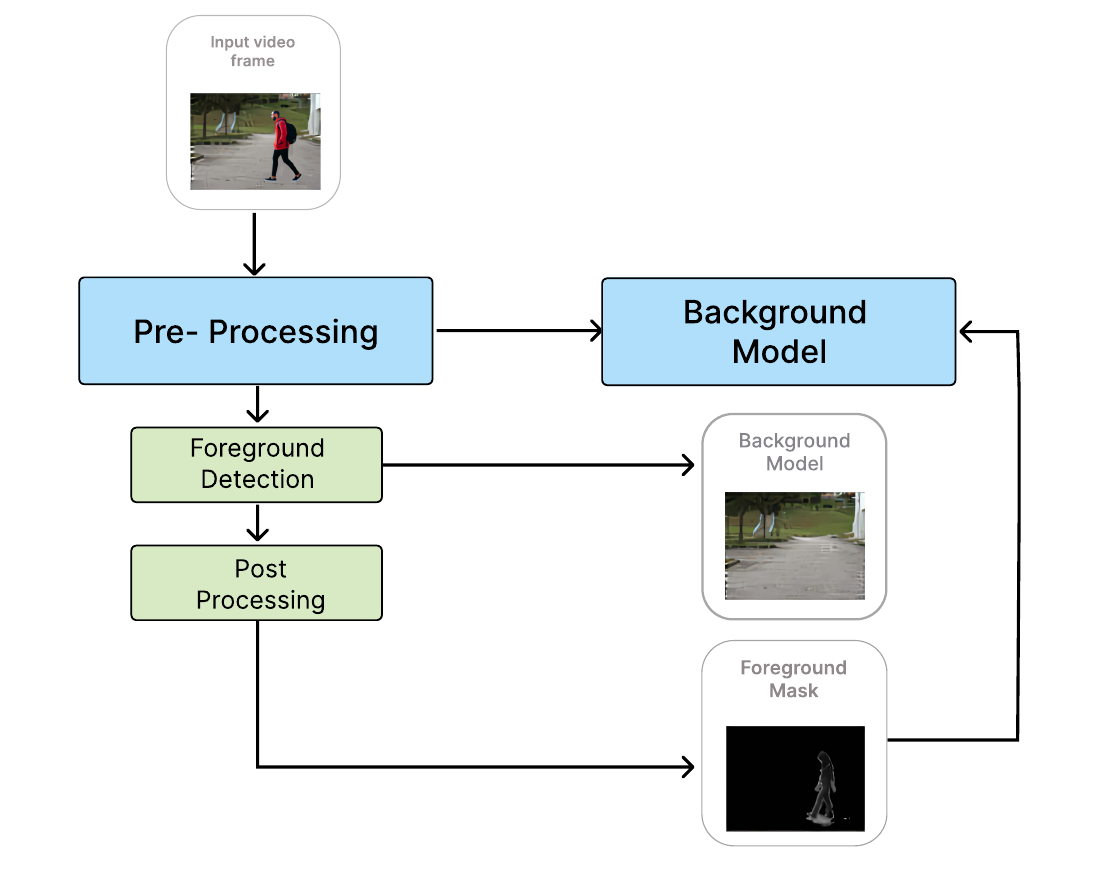}
    \caption{MOG2 background subtraction algorithm process}
    \label{fig:mog1}
\end{figure}

This image \hyperref[fig:mog1]{Figure~\ref*{fig:mog1} } illustrates the  background subtraction  process. The process starts with preparing the input video frames, then moves on to background modeling to identify the foreground. The background model is constantly adjusted, and post-processing is used to enhance the accuracy of detecting moving objects by refining the foreground mask.

\subsection{Related Works}
These papers subjected on object detecting and count it that concluding issues for an electoral 
controversial solution. 
A computer vision-based method for person counting and monitoring entrances and exits inside a scene 
is presented in the publication "\textit{Robust Background Subtraction Based Person's Counting from 
Overhead View.}" After testing a number of background reduction techniques with an overhead camera, 
the study concludes that the Mixture of Gaussian (MoG) algorithm works best for top-down person 
counting. This method tracks people entering and leaving the area by using a predefined virtual zone. 
Based on the ground truth of the created dataset with the abridgment of limited subject in constrained 
environment, the algorithm achieves high accuracy: 98\% for entries and exits within the virtual zone 
and 98\% for human counting. The MoG approach demonstrates strong noise rejection and flexibility, 
with room for further development to identify additional users, ascertain their directions of movement, 
and incorporate classification techniques for more extensive datasets \cite{20}.

According to Uke and Thool \cite{16}, the paper introduces a smart counting and vehicle detection system 
created using Visual C++ and also Intel's OpenCV, utilizing OpenCV image enhancement kits for 
implementation. In today's busy traffic scenarios, accurately monitoring vehicles is crucial for tasks 
such as highway oversight, traffic flow planning, and toll collection. What's intriguing about this system 
is 
its reliance on Computer Vision techniques, ensuring seamless installation and adaptable 
functionality. It's cost-effective, portable, and proficient at identifying various vehicle types—ranging 
from light vehicles to heavy ones and motorcycles. Rigorous testing on a standard laptop validated its 
efficiency in real-time vehicle counting using methods like background subtraction and image filtering. 
Moreover, its ability to count vehicles from both live feeds and pre-recorded videos positions it as a 
potential game-changer in optimizing traffic management systems. 

About the paper to count passengers in a bus, the application of OpenCV’s MOG2 algorithm for 
accurate passenger counting in a ticket-less bus service during a housing fair is described in depth in 
the paper. In order to track passengers, it combines shape detection and picture analysis, focusing on 
the number of people entering and leaving various bus stops. Route optimization and statistical data 
collection are the main goals in order to prepare for future events. This study demonstrates how 
computer vision techniques, particularly the MOG2 algorithm with OpenCV, can be used to accurately 
and economically count the number of individuals in situations when more traditional methods fall short 
\cite{21}. Experimental results were demonstrated using real-time video captured from a 
single camera, confirming its practical suitability and performance. 

Sabancı et al. \cite{23} argues the vital junction of computer-human interaction and image processing, 
particularly in systems employing cameras to count in library. The writers stress how crucial object 
identification is to improving computer capabilities—particularly in dynamic contexts. They discuss 
problems including the intricacies of real-time processing and provide fixes like shadow detection and 
background reduction. Notably, the integration of Internet of Things (IoT) concepts, utilizing the 
Thingspeak platform, is introduced for efficient data transfer. After that, the paper turns to a practical 
application and presents an autonomous system designed specifically for people counting in spaces. 
Originally created to help students discover study areas during exam weeks, this system demonstrates 
adaptability for more general applications including sensor data transfer. The project's potential 
significance is emphasized by the authors, who also highlight how it might help overcome certain obstacles and show how scalable it is for a range of sensor-related demands.

In order to detect criminal activity in public areas, the article “\textit{Motion and Object Recognition for Crime 
Prediction and Forecasting: A Review}” suggests using an algorithm to create an automated surveillance 
system \cite{24}. The technology uses motion detection and object classification based on machine learning with 
the goal of improving public safety. Techniques for feature extraction, detection, and motion are 
essential for spotting patterns that point to aggressive or strange behavior. The project uses OpenCV for 
background subtraction including the MOG2 algorithm, video feed recording, and Python and 
Computer Vision are used to implement motion detection. By analyzing CCTV camera footage, the 
system can differentiate between routine and questionable activity in public areas. Important procedures 
including dataset preparation, pre-processing, model training, and evaluation are part of the detection 
process. By assisting law enforcement authorities in identifying patterns and contributing causes to 
violent behavior, the goal is to prevent violent situations through early intervention. The system exhibits 
the varied application of technology in crime prediction and prevention by directing police patrols 
towards potential criminal activities through the implementation of machine learning algorithms. 

“\textit{Pedestrian flow counter using image processing}” presents a human counting system that counts people 
reliably in a variety of settings, including retail stores, malls, and ATMs \cite{25}. It does this by using Image 
Processing techniques. The system uses a WiFi feed from an IP CCTV camera overhead that is 
processed by Python's OpenCV module. To reduce noise and improve blobs, morphological treatments 
are conducted frame-by-frame to the live video input. Blob coordinates are extracted using the contours 
approach, which enables efficient tracking and counting inside predetermined frame zones. The report 
goes into great detail about how lighting, camera angle, camera height, and range of vision affect 
counting accuracy. For effective operation, the suggested People Counter Combines Blob Tracking, 
Area Contours, and Background Subtraction. Morphological changes are used to ensure correct 
background subtraction. Important environmental issues include illumination and camera angle. 
Because image processing requires a lot of computing power, the paper also compares the performance 
of an expensive GPU-powered i5 processor with other onboard processors .
\vspace{.2cm}
\section{Methodology}
\label{sec:phase}
A mixed-method approach is used in the study, which combines quantitative examination of computer vision-based automated vote counting with qualitative analysis of electoral transparency issues. The study evaluates the accuracy and effectiveness of real-time voter counting in the 2024 elections using the Mog2 algorithm and libraries such as CVZone and OpenCV respectively.A mixed-method approach is used in the study, which combines quantitative examination of computer vision-based automated vote counting with qualitative analysis of electoral transparency issues. The study evaluates the accuracy and effectiveness of real-time voter counting in the 2024 elections using the Mog2 algorithm and libraries such as CVZone and OpenCV respectively.

\begin{figure}[H]
    \centering
    \includegraphics[width=0.8\textwidth]{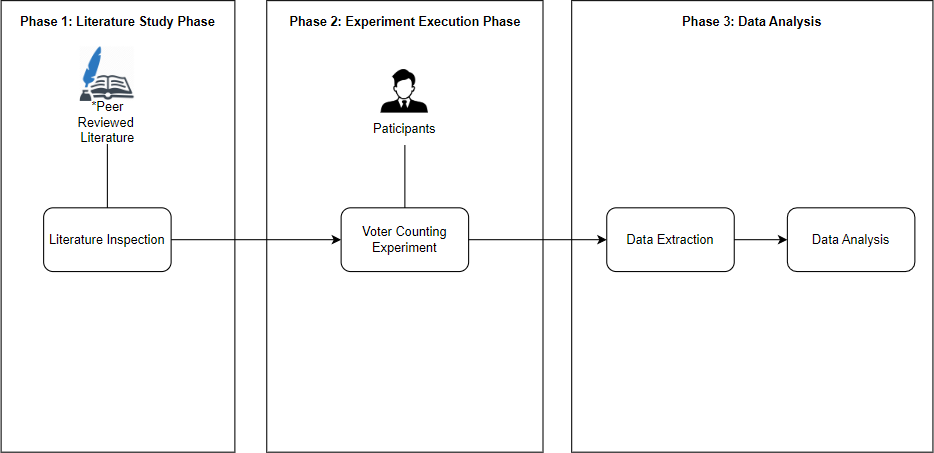}
    \caption{All The Phases of the Evaluation Process.}
    \label{fig:phases}
\end{figure}

\subsection{Phase 1: Literature Review}
Examining previous studies on computer vision integration, image processing techniques, and automated voter counting systems in election processes is the main goal of the literature review phase. This entails combining various studies, evaluating the advantages and disadvantages, looking at case studies from around the world, and seeing patterns to guide the development of transparent, reliable, and accurate
voting systems.
\subsubsection{Comprehensive Literature Search}
Conduct a meticulous and systematic review of a
diverse range of academic databases, journals, conference proceedings, and authoritative
publications to assemble a comprehensive collection of relevant literature.
\subsubsection{Thematic Categorization}
Organize the gathered literature into distinct themes,
encompassing topics such as advanced image processing algorithms, the utilization of computer
vision in voting systems, and the evolution of automated voter counting mechanisms.
\subsubsection{Critical Synthesis}
 Perform a critical synthesis of the identified literature, emphasizing the
strengths, weaknesses, trends, and gaps in existing methodologies. This synthesis will provide a
nuanced understanding of the current state of automated voter counting systems.
\subsubsection{Cross-disciplinary Exploration}
Incorporate interdisciplinary research to gain a
comprehensive knowledge that encompasses technology, ethics, and government.

\subsection{Phase 2: Voter Counting Experiment}
The second phase endeavors to conduct a meticulously designed experiment emulating a real-world 
voting scenario. This involves recording the voting process and implementing cutting-edge computer 
vision techniques for automated voter counting.
\subsubsection{Experimental Design}
Establish a controlled voting environment, ensuring its authenticity 
and representatives of actual voting scenarios. Consider factors such as participant behavior, 
spatial dynamics, and potential challenges. 
\subsubsection{Placement of Device}
The device should be placed on a suitable clear view area in front of 
booth entry.
\subsubsection{High-Resolution Recording}
 Employ state-of-the-art video recording equipment to capture 
the entire voting process with meticulous attention to detail, ensuring clarity in individual 
movements and interactions. It requires high level device for capturing the modern included 
features.
\subsubsection{Advanced Computer Vision Implementation}
This work employs OpenCV's MOG2 algorithm, which is essential for automated voter counting, to detect people accurately through efficient background subtraction. The incorporation of CVZone improves dynamic motion tracking, and the generation of dynamic lines boosts entry/exit demarcation, lowering miscounts and improving system accuracy.

\begin{figure}[H]
    \centering
    \includegraphics[width=0.7\textwidth]{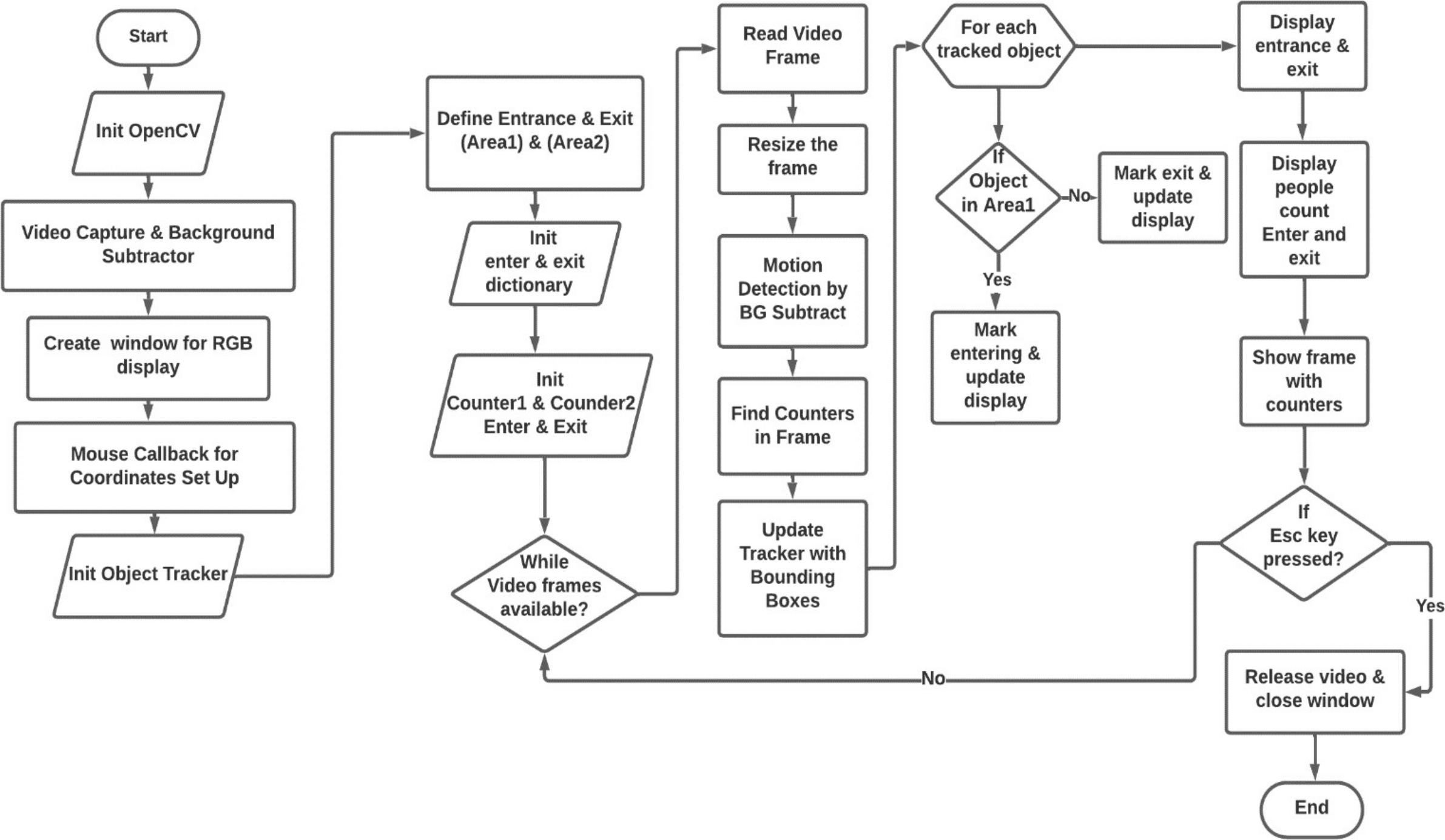}
    \caption{A comprehensive flow chart diagram.}
    \label{fig:flowchart}
\end{figure}

\subsubsection{Monitoring Environment Variables}
Keeping an eye on how environmental factors 
affect the precision of the automated counting system when voting in order to conduct a thorough 
investigation. Such as MOG2 on contradiction to selection of shadow.

\subsubsection{Quantitative Data Collection}
Systematically extract quantitative data encompassing the number of individuals entering and exiting the voting area throughout the recorded session.

\subsection{Phase 3: Data Analysis}
The third phase involves a rigorous analysis of the extracted data, comparing automated voter counting
results with manually derived ground truth to evaluate the accuracy, efficiency, and reliability of the
implemented system.

\subsubsection{Data Extraction and Transformation}
Extract pertinent data points from the recorded
video, employing advanced data transformation techniques to ensure compatibility and
coherence.

\subsubsection{Ground Truth Establishment}
Establish a robust ground truth by manually counting
individuals in the recorded video, serving as a benchmark for evaluating the accuracy of the
automated voter counting system.

\subsubsection{Quantitative and Qualitative Analysis}
In the quantitative analysis phase, the evaluation
of the automated voter counting system will extend beyond traditional metrics like accuracy and
precision. The inclusion of the F1 score adds a layer of sophistication to the assessment by considering both precision and recall simultaneously \cite{22}. This is crucial, especially in scenarios where imbalanced datasets or equitable consideration of false positives and false negatives are paramount. The F1 score is a metric often used in machine learning and
statistics to evaluate the effectiveness of a classification model, especially when dealing with
imbalanced datasets or confusion matrices. It offers a balance between these two measurements and is the harmonic mean of recall and precision. Precision (also called positive predictive value) can be calculated by dividing the total number of positive predictions by the number of real positive forecasts. It gauges how accurate positive forecasts are . The mathematical process of positive forecasts \cite{28} are:

\begin{equation}
\text{Precision} = \frac{TP}{TP + FP}
 \label{eq:precision}
 \end{equation}

Recall (also called sensitivity or true positive rate) is the number of true positive predictions divided by the total number of actual positive instances. It measures the ability of the model to capture all the positive instances. The mathematical process of Recall \cite{29} is:

\begin{equation}
\text{Recall} = \frac{TP}{TP + FN}
 \label{eq:recall}
 \end{equation}

The harmonic mean of recall and precision is then used to compute the F1 score. The mathematical process of F1 score \cite{30} is:

\begin{equation}
F1 \text{ Score} = 2 \times \frac{\text{Precision} \times \text{Recall}}{\text{Precision} + \text{Recall}}
 \label{eq:f1score}
 \end{equation}

The F1 score ranges from 0 to 1, where 1 indicates perfect precision and recall, and 0 indicates the worst performance. It is a useful metric when there is an uneven class distribution or when both false positives and false negatives are important considerations for the task at hand  \cite{22}.

\subsection{Statistical Significance Testing}
 Apply robust statistical methods to assess the significance 
of disparities between automated and manual counting results, providing a scientific basis for the 
system's efficacy.

\subsection{Comprehensive Result Presentation}
Present findings comprehensively, incorporating visualizations, statistical summaries, and insights into the strengths and weaknesses of the 
automated voter counting system. Propose areas for refinement and potential avenues for future 
research.

\section{Results}
The results section presents the findings of the experiment. It demonstrate how the automated voter counting system performed under various conditions, with a particular focus on accuracy and efficiency.

\subsection{Overview}
The automated voter counting system was developed using a custom dataset, created with the
collaboration of friends, simulating a voting environment. The dataset was generated by recording
individuals using a camera within this simulated voting setting. The following processes were applied to
the recorded data using OpenCV, CVZone, and the MOG2 algorithm.

\subsection{Data Collection}
The custom dataset was systematically collected, capturing individuals participating in the voting simulation. Utilizing OpenCV, CVZone, and the MOG2 algorithm, the recorded video frames were processed to identify and count individuals entering and exiting the voting area.

\begin{figure}[H]
    \centering
    \includegraphics[width=0.7\textwidth]{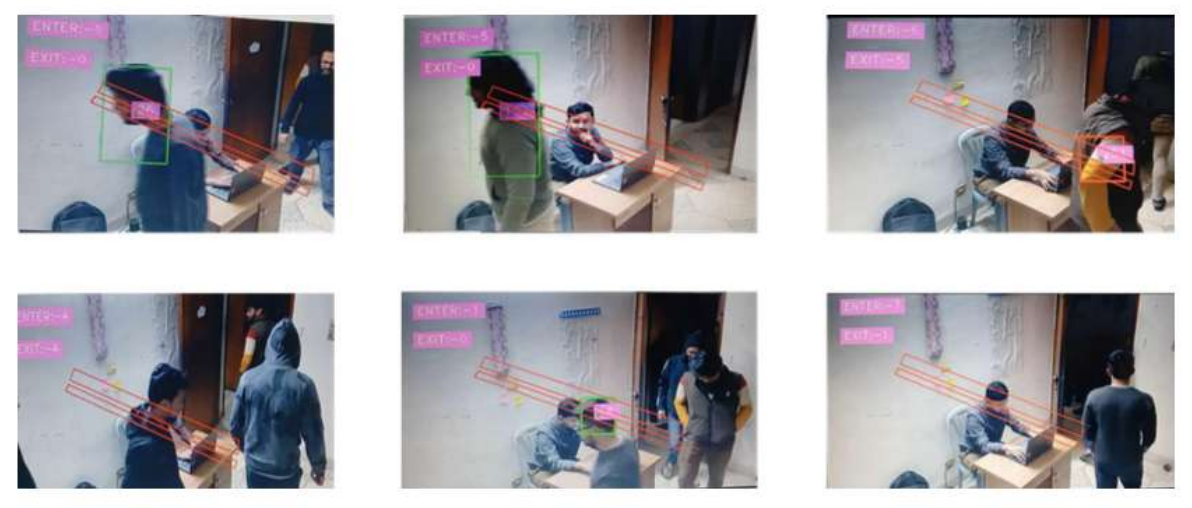}
    \caption{Dataset for voter counting.}
    \label{fig:dataset}
\end{figure}

This image \hyperref[fig:dataset]{Figure~\ref*{fig:dataset} } provides a visual representation of the data collection process in the voting simulation, employing advanced computer vision techniques. Utilizing a custom dataset meticulously gathered during the simulation, the image showcases the successful application of OpenCV, CVZone.

\subsection{Analysis of Results}
The effectiveness of the automated voter counting system was assessed by comparing its output with manually verified ground truth data. As shown in Table 1, the system performed well across multiple test scenarios involving different numbers of participants. Each scenario measured the system’s ability to accurately count both entries and exits within the polling area.

\begin{table}[ht]
    \centering
    \caption{Experimental Results of Automated Voter Counting System}
    \begin{tabular}{|c|c|c|c|}
        \hline
        \textbf{Number of People} & \textbf{Entering Accuracy} & \textbf{Leaving Accuracy} & \textbf{Average Accuracy} \\ \hline
        2  & 100\% & 100\% & 100\%  \\ \hline
        4  & 100\% & 100\% & 100\%  \\ \hline
        10 & 100\% & 98.3\% & 99.15\% \\ \hline
    \end{tabular}
    \label{tab:experimental-results}
\end{table}

This table \ref {tab:experimental-results} delves into the specifics of system performance, presenting the accuracy in counting people
entering and exiting in various scenarios. The consistent accuracy of 99.15]\%  across different situations highlights the reliability of the system. The overall accuracy remains robust at 99.15 \%  indicating the system's effectiveness in accurately detecting and counting individuals in the dynamic voting environment.

Let's define the following:

\textbf{For entering:}
\begin{itemize}
    \item True Positives (TP\_enter): 30 (people who both entered and were counted)
    \item False Positives (FP\_enter): 0 (people who were counted as entering but did not enter)
    \item False Negatives (FN\_enter): 0 (people who entered but were not counted)
\end{itemize}

\[
\text{Precision} = \frac{TP_{\text{enter}}}{TP_{\text{enter}} + FP_{\text{enter}}} = \frac{30}{30 + 0} = 1
\]
\[
\text{Recall} = \frac{TP_{\text{enter}}}{TP_{\text{enter}} + FN_{\text{enter}}} = \frac{30}{30 + 0} = 1
\]
\[
F1_{\text{enter}} = 2 \times \frac{\text{Precision} \times \text{Recall}}{\text{Precision} + \text{Recall}} = 1 = 100\%
\]

\textbf{For exiting:}
\begin{itemize}
    \item True Positives (TP\_exit): 29 (people who both exited and were counted)
    \item False Positives (FP\_exit): 1 (people who were counted as exiting but did not exit)
    \item False Negatives (FN\_exit): 1 (people who exited but were not counted)
\end{itemize}

\[
\text{Precision} = \frac{TP_{\text{exit}}}{TP_{\text{exit}} + FP_{\text{exit}}} = \frac{29}{30}
\]
\[
\text{Recall} = \frac{TP_{\text{exit}}}{TP_{\text{exit}} + FN_{\text{exit}}} = \frac{29}{30}
\]
\[
F1_{\text{exit}} = 2 \times \frac{\text{Precision} \times \text{Recall}}{\text{Precision} + \text{Recall}} = 0.983 = 98.3\%
\]

Next step is to calculate precision, recall, and F1 score for both entering and exiting, and then find the average.

\[
\text{Average F1 Score:}
\]
\[
\text{Average F1Score} = \frac{F1_{\text{enter}} + F1_{\text{exit}}}{2} = \frac{1 + 0.983}{2} = 0.9915 = 99.15\%
\]

\begin{table}[h!]
\centering
\caption{Overview of People Counting Accuracy}
\begin{tabular}{|>{\raggedright\arraybackslash}m{4cm}|>{\centering\arraybackslash}m{3cm}|>{\raggedright\arraybackslash}m{7cm}|}
\hline
\textbf{Activity} & \textbf{Counting Result} & \textbf{Details} \\ \hline
People Enter & 100\% & Perfect accuracy in counting people entering the voting environment. \\ \hline
People Exit & 98.3\% & High accuracy of 98.3\% in counting people exiting the voting area. \\ \hline
Overall Counting Average & 99.15\% & Impressive accuracy of 99.15\% when considering both entry and exit scenarios. \\ \hline
\end{tabular}
\label{tab:accuracy}
\end{table}

This table \ref{tab:accuracy} outlines the accuracy of the automated voter counting system in different scenarios. Notably,
the system achieved perfect accuracy in counting individuals entering the voting environment,
showcasing its precision. Moreover, the high accuracy of 98.3\% in counting people exiting reflects the system's effectiveness. The overall counting average reached an impressive 99.15\%, emphasizing the reliability of the system in monitoring both entry and exit scenarios.

\begin{table}[h!]
\centering
\caption{Implementation Details and Findings}
\begin{tabular}{|>{\raggedright\arraybackslash}m{4cm}|>{\raggedright\arraybackslash}m{5cm}|>{\raggedright\arraybackslash}m{7cm}|}
\hline
\textbf{Implementation} & \textbf{Details} & \textbf{Findings} \\ \hline
Libraries Used & CVZone and OpenCV & Utilized for real-time video frame analysis, object detection, and tracking individuals in the voting environment. \\ \hline
Algorithm Used & MOG2 & Applied for background subtraction, enhancing the accuracy of object identification. \\ \hline
Dataset & Created a voting environment dataset & Used to monitor people entering and exiting, simulating a real-world voting scenario. \\ \hline
Data Collection & Quantitative data on individuals & Collected information on the number of people entering and exiting based on video recordings. \\ \hline
Accuracy Achieved & 99.15\% & Successfully detected and counted individuals in the voting environment. \\ \hline
\end{tabular}
\label{tab:findings}
\end{table}

This table \ref{tab:findings} provides insights into the implementation details of the automated voter counting system. The
use of CVZone and OpenCV libraries, coupled with the MOG2 algorithm, enabled real-time video frame analysis, object detection, and tracking of individuals. The creation of a voting environment dataset facilitated the simulation of real-world scenarios, contributing to the systematic collection of quantitative data. The system achieved an impressive accuracy of 99.15\%, successfully detecting and counting individuals during the voting process.

\subsection{Key Findings}
Accuracy: The automated voter counting system demonstrated an accuracy rate of 99.15\%, showcasing its effectiveness in accurately detecting and counting individuals within the voting environment.
The utilization of the custom dataset, combined with the implementation of OpenCV, CVZone, and the MOG2 algorithm, yielded a robust automated voter counting system. The system's accuracy of 99.15\% and additional performance metrics emphasized its reliability in accurately counting individuals within a simulated voting environment. The findings from this analysis contribute to the advancement of automated systems for enhancing the efficiency and transparency of the electoral process.

 \section{Discussion}
 \label{sec:discussion}
The automated system, in contrast to the traditional manual vote counting manner greatly increases the accuracy and efficiency levels. Manual counting is also error-prone, and unexpected discrepancies can arise between the results reported by nominated candidates. In comparison, an automated system can produce exact and up-to-date results swiftly without much error lead to a viable solution of enhancing the elections transparency as well as credibility.

Although accuracy of the system high, there are limitations were seen during experiment. Although, the system was somewhat sensitive to light changes and occlusions (where one voter would block another from camera view). This suggests that the system could be improved to more effectively navigate complex real-world scenarios.

 \section{Conclusion}
 \label{sec:conclusion}
To avoid the disadvantages of manual vote process image processing based automated system was introduced.
Some computer vision and machine learning tools such as OpenCV, CVZone, MOG2 algorithm have been used to design the project. This
rates over which reduce election efficiency dramatically. This approach has the potential to do just that, with an accuracy rate of 99.15\%
port, and finally in places with weak electoral process trust such as Bangladesh. By offering precise input and output including real-time results, the approach can assuage fears about vote tampering and delays justification that tend to undermine corrupt processes. Further work may focus on integrating bio-metric or another authentication method of identity verification.
Biometrics such as facial recognition system to make the system even more secure and reliable. As well as integration of database cloud computing where each registered citizen get detected uniquely. Additionally, to ensure the wide acceptance and success of these technologies, raising public awareness on data practices create special benefits of automated voter counting by extension. The legitimacy of democracy in eyes, is fixed and the mechanical components of vote counting are refined.

\section*{CRediT Authorship Contribution Statement}
\textbf{Mir Faris:} Conceptualization, Investigation, Writing Original Draft, Research administration. \& Editing,  Resources.
\textbf{Syeda Aynul Karim:} Methodology, Investigation, Research administration., Writing Original Draft,\newline
\textbf{Md Juniadul Islam:} Methodology, Investigation, Writing Original Draft, Writing - Review \& Editing,  Resources.

\section*{Declaration of interest’s statement}

There are no competing interests amongst the authors of this article, they declared. As far as the work we have submitted is concerned, we thus declare that we have no competing interests or affiliations.

\bibliographystyle{elsarticle-num} 
\bibliography{ref_Unmarked}

% \include{main2}





% \end{thebibliography}
\end{document}